\documentclass{article}

\usepackage{microtype}
\usepackage{graphicx}
\usepackage{subfigure}
\usepackage{booktabs} 
\usepackage{times}

\usepackage{soul}
\usepackage{url}
\usepackage[utf8]{inputenc}
\usepackage[small]{caption}
\usepackage{amsmath,amssymb}
\usepackage{booktabs}
\usepackage{amsthm}
\usepackage{amssymb}

\usepackage{algorithm}
\usepackage[noend]{algorithmic}
\usepackage{enumitem}

\DeclareMathOperator{\E}{\mathbb{E}}

\usepackage{hyperref}

\newcommand{\eg}{\emph{e.g.}}
\newcommand{\etc}{\emph{etc}}
\newcommand{\ie}{\emph{i.e.}}

\usepackage[accepted]{icml2020}

\icmltitlerunning{Neural-Grammar-Symbolic}

\begin{document}

\twocolumn[
\icmltitle{Closed Loop Neural-Symbolic Learning via \\ Integrating Neural Perception, Grammar Parsing, and Symbolic Reasoning}



\icmlsetsymbol{equal}{*}

\begin{icmlauthorlist}
\icmlauthor{Qing Li}{ucla}
\icmlauthor{Siyuan Huang}{ucla}
\icmlauthor{Yining Hong}{ucla}
\icmlauthor{Yixin Chen}{ucla}
\icmlauthor{Ying Nian Wu}{ucla}
\icmlauthor{Song-Chun Zhu}{ucla}
\end{icmlauthorlist}

\icmlaffiliation{ucla}{ University of California, Los Angeles, USA}

\icmlcorrespondingauthor{Qing Li}{liqing@ucla.edu}

\icmlkeywords{Machine Learning, ICML}

\vskip 0.3in
]



\printAffiliationsAndNotice{}  

\begin{abstract}
The goal of neural-symbolic computation is to integrate the connectionist and symbolist paradigms. Prior methods learn the neural-symbolic models using reinforcement learning (RL) approaches, which ignore the error propagation in the symbolic reasoning module and thus converge slowly with sparse rewards. In this paper, we address these issues and close the loop of neural-symbolic learning by (1) introducing the \textbf{grammar} model as a \textit{symbolic prior} to bridge neural perception and symbolic reasoning, and (2) proposing a novel \textbf{back-search} algorithm which mimics the top-down human-like learning procedure to propagate the error through the symbolic reasoning module efficiently. We further interpret the proposed learning framework as maximum likelihood estimation using Markov chain Monte Carlo sampling and the back-search algorithm as a Metropolis-Hastings sampler. The experiments are conducted on two weakly-supervised neural-symbolic tasks: (1) handwritten formula recognition on the newly introduced HWF dataset; (2) visual question answering on the CLEVR dataset. The results show that our approach significantly outperforms the RL methods in terms of performance, converging speed, and data efficiency. Our code and data are released at \url{https://liqing-ustc.github.io/NGS}.
\end{abstract}

\section{Introduction}
\begin{figure}[ht]
\begin{center}
\centerline{\includegraphics[width=\columnwidth]{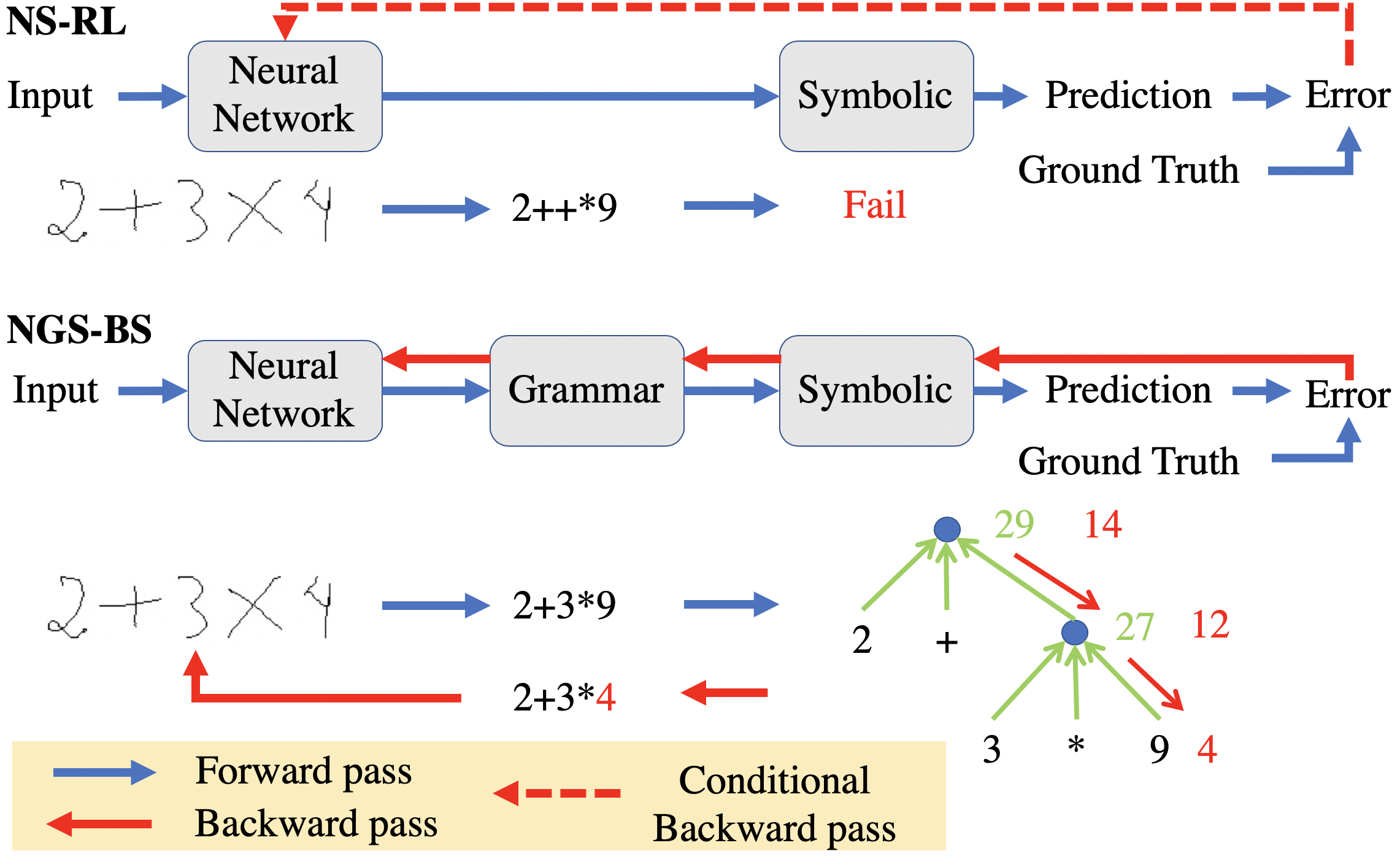}}
\caption{Comparison between the original neural-symbolic model learned by REINFORCE (NS-RL) and the proposed neural-grammar-symbolic model learned by back-search (NGS-BS). In NS-RL, the neural network predicts an invalid formula, causing a failure in the symbolic reasoning module. There is no backward pass in this example since it generates zero reward. In contrast, NGS-BS predicts a valid formula and searches a correction for its prediction. The neural network is updated using this correction as the pseudo label. }\label{fig_RL_vs_BS}
\end{center}
\end{figure}
Integrating robust connectionist learning and sound symbolic reasoning is a key challenge in modern Artificial Intelligence. Deep neural networks~\cite{lecun2015deep,lecun1995convolutional,hochreiter1997long} provide us powerful and flexible representation learning that has achieved state-of-the-art performances  across a variety of AI tasks such as image classification~\cite{krizhevsky2012imagenet,szegedy2015going,he2016deep}, machine translation~\cite{sutskever2014sequence}, and speech recognition~\cite{graves2013speech}. However, it turns out that many aspects of human cognition, such as systematic compositionality and generalization~\cite{fodor1988connectionism,marcus1998rethinking,fodor2002compositionality,calvo2014architecture,marcus2018algebraic,Lake2018Generalization}, cannot be captured by neural networks. On the other hand, symbolic reasoning supports strong abstraction and generalization but is fragile and inflexible. 
Consequently, many methods have focused on building neural-symbolic models to combine the best of deep representation learning and symbolic reasoning \cite{sun1994integrating, garcez2008neural, bader2009extracting, besold2017neural, yi2018neural}.

Recently, this neural-symbolic paradigm has been extensively explored in the tasks of the visual question answering (VQA) \cite{yi2018neural, vedantam2019probabilistic, mao2019neuro}, vision-language navigation \cite{anderson2018vision,fried2018speaker}, embodied question answering \cite{das2018embodied,das2018neural}, and semantic parsing \cite{liang2016neural, yin2018structvae}, often with weak supervision. Concretely, for these tasks, neural networks are used to map raw signals (images/questions/instructions) to symbolic representations (scenes/programs/actions), which are then used to perform symbolic reasoning/execution to generate final outputs. Weak supervision in these tasks usually provides pairs of raw inputs and final outputs, with intermediate symbolic representations unobserved. Since symbolic reasoning is non-differentiable, previous methods usually learn the neural-symbolic models by policy gradient methods like REINFORCE. The policy gradient methods generate samples and update the policy based on the generated samples that happen to hit high cumulative rewards. No efforts are made to improve each generated sample to increase its cumulative reward.  Thus the learning has been proved to be time-consuming because it requires generating a large number of samples over a large latent space of symbolic representations with sparse rewards, in the hope that some samples may be lucky enough to hit high rewards so that such lucky samples can be utilized for updating the policy. As a result, policy gradients methods converge slowly or even fail to converge without pre-training the neural networks on fully-supervised data.

To model the recursive compositionality in a sequence of symbols, we introduce the \textbf{grammar} model to bridge neural perception and symbolic reasoning. The structured symbolic representation often exhibits compositional and recursive properties over individual symbols in it. Correspondingly, the grammar models encode \textit{symbolic prior} about composition rules, thus can dramatically reduce the solution space by parsing the sequence of symbols into valid sentences. For example, in the handwritten formula recognition problem, the grammar model ensures that the predicted formula is always valid,  as shown in \autoref{fig_RL_vs_BS}. 

To make the neural-symbolic learning more efficient, we propose a novel \textbf{back-search} strategy which mimics human's ability to learn from failures via abductive reasoning~\cite{magnani2009abductive,zhou2019abductive}. Specifically, the back-search algorithm propagates the error from the root node to the leaf nodes in the reasoning tree and finds the most probable \textit{correction} that can generate the desired output. The correction is further used as a pseudo label for training the neural network. \autoref{fig_RL_vs_BS} shows an exemplar backward pass of the back-search algorithm. We argue that the back-search algorithm makes a first step towards closing the learning loop by propagating the error through the non-differentiable grammar parsing and symbolic reasoning modules. We also show that the proposed multi-step back-search algorithm can serve as a Metropolis-Hastings sampler which samples the posterior distribution of the symbolic representations in the maximum likelihood estimation in ~\autoref{sec:mBS}.

We conduct experiments on two weakly-supervised neural-symbolic tasks: (1) handwritten formula recognition on the newly introduced HWF dataset (\underline{H}and-\underline{W}ritten \underline{F}ormula), where the input image and the formula result are given during training, while the formula is hidden; (2) visual question answering on the CLEVR dataset. The question, image, and answer are given, while the functional program generated by the question is hidden. The evaluation results show that the proposed Neural-Grammar-Symbolic (NGS) model with back-search significantly outperforms the baselines in terms of performance, convergence speed, and data efficiency. The ablative experiments also demonstrate the efficacy of the multi-step back-search algorithm and the incorporation of grammar in the neural-symbolic model. 



\section{Related Work}

\noindent \textbf{Neural-symbolic Integration.}
Researchers have proposed to combine statistical learning and symbolic reasoning in the AI community, with pioneer works devoted to different aspects including representation learning and reasoning~\cite{sun1994integrating, garcez2008neural,manhaeve2018deepproblog}, abductive learning~\cite{dai2017combining,dai2019bridging,zhou2019abductive}, knowledge abstraction~\cite{hinton2006fast, bader2009extracting}, knowledge transfer~\cite{falkenhainer1989structure, yang2009heterogeneous}, \etc. Recent research shifts the focus to the application of neural-symbolic integration, where a large amount of heterogeneous data and knowledge descriptions are needed, such as neural-symbolic VQA~\cite{yi2018neural, vedantam2019probabilistic,mao2019neuro,gan2017vqs}, semantic parsing in Natural Language Processing (NLP)~\cite{liang2016neural, yin2018structvae}, math word problem~\cite{lample2019deep,lee2019mathematical} and program synthesis~\cite{evans2018learning, kalyan2018neural, manhaeve2018deepproblog}. Different from previous methods, the proposed NGS model considers the compositionality and recursivity in natural sequences of symbols and brings together the neural perception and symbolic reasoning module with a grammar model.

\noindent \textbf{Grammar Model.}
Grammar model has been adopted in various tasks for its advantage in modeling compositional and recursive structures, like image parsing~\cite{tu2005image,han2005bottom,zhu2007stochastic,zhao2011image,friesen2018submodular}, video parsing~\cite{gupta2009understanding,qi2018generalized,qi2020generalized}, scene understanding~\cite{huang2018holistic,huang2018cooperative,qi2018human,jiang2018configurable,chen2019holistic++}, and task planning~\cite{xu2018unsupervised}. By integrating the grammar into the neural-symbolic task as a symbolic prior for the first time, the grammar model ensures the desired dependencies and structures for the symbol sequence and generates valid sentences for symbolic reasoning. Furthermore, it improves the learning efficiency significantly by shrinking the search space with the back-search algorithm. 

\noindent \textbf{Policy Gradient.}
Policy gradient methods like REINFORCE~\cite{williams1992simple} are the most commonly used algorithm for the neural-symbolic tasks to connect the learning gap between neural networks and symbolic reasoning~\cite{mascharka2018transparency,mao2019neuro,andreas2017modular,das2018neural,bunel2018leveraging,guu2017language}. However, original REINFORCE algorithm suffers from large sample estimate variance, sparse rewards from cold start and exploitation-exploration dilemma, which lead to unstable learning dynamics and poor data efficiency. Many papers propose to tackle this problem~\cite{liang2016neural, guu2017language, Liang2018MemoryAP, wang2018mathdqn, agarwal2019learning}. Specifically, \citet{liang2016neural} uses iterative maximum likelihood to find pseudo-gold symbolic representations, and then add these representations to the REINFORCE training set. \citet{guu2017language} combines the systematic beam search
employed in maximum marginal likelihood with the
greedy randomized exploration of REINFORCE. \citet{Liang2018MemoryAP} proposes Memory Augmented Policy Optimization (MAPO) to express the expected return objective as a weighted sum of an expectation over the high-reward history trajectories, and a separate expectation over new trajectories. Although utilizing positive representations from either beam search or past training process, these methods still cannot learn from negative samples and thus fail to explore the solution space efficiently. On the contrary, we propose to diagnose and correct the negative samples through the back-search algorithm under the constraint of grammar and symbolic reasoning rules. Intuitively speaking, the proposed back-search algorithm traverses around the negative sample and find a nearby positive sample to help the training.

\section{Neural-Grammar-Symbolic Model (NGS)} \label{sec_method}
In this section, we will first describe the inference and learning algorithms of the proposed neural-grammar-symbolic (NGS) model. Then we provide an interpretation of our model based on maximum likelihood estimation (MLE) and draw the connection between the proposed back-search algorithm and Metropolis-Hastings sampler. We further introduce the task-specific designs in \autoref{sec:experiments}.


\subsection{Inference}
In a neural-symbolic system, let $x$ be the input (\eg  an image or question), $z$  be the hidden symbolic representation, and $y$  be the desired output inferred by $z$. The proposed NGS model combines neural perception, grammar parsing, and symbolic reasoning modules efficiently to perform the inference.


\noindent \textbf{Neural Perception}. The neural network is used as a perception module which maps the high-dimensional input $x$ to a normalized probability distribution of the hidden symbolic representation $z$:
\begin{align} \label{eq_neural}
p_\theta(z|x) &= softmax(\phi_\theta(z,x)) \\
&=\frac{\exp(\phi_\theta(z,x))}{\sum_{z'} \exp(\phi_\theta(z',x))}, 
\end{align}
where $\phi_\theta(z,x)$ is a scoring function or a negative energy function represented by a neural network with parameters $\theta$. 

\noindent \textbf{Grammar Parsing}. Take $z$ as a sequence of individual symbols: $z = (z_1, z_2, ..., z_l), z_i \in \Sigma$, where $\Sigma$ denotes the vocabulary of possible symbols. The neural network is powerful at modeling the mapping between $x$ and $z$, but the recursive  compositionality among the individual symbols $z_i$ is not well captured. Grammar is a natural choice to tackle this problem by modeling the compositional properties in sequence data.

Take the \textit{context-free grammar} (CFG) as an example. In formal language theory, a CFG is a type of formal grammar containing a set of production rules
that describe all possible sentences in a given formal language. Specifically, a context-free grammar $G$ in Chomsky Normal Form is defined by a 4-tuple $G = (V, \Sigma, R, S)$, where
\begin{itemize}
    \setlength\itemsep{0em}
    \item $V$ is a finite set of non-terminal symbols that can be
replaced by/expanded to a sequence of symbols.
    \item $\Sigma$ is a finite set of terminal symbols that represent actual words in a language, which cannot be further expanded. Here $\Sigma$ is the vocabulary of possible symbols.
    \item $R$ is a finite set of production rules describing the replacement of symbols, typically of the form $A \to BC$ or $A \to \alpha$, where $A, B, C \in V$ and $\alpha \in \Sigma$.  A production rule replaces the left-hand side non-terminal symbols by the right-hand side expression. For example, $A \to BC|\alpha$ means that $A$ can be replaced by either $BC$ or $\alpha$.
    \item $S \in V$ is the start symbol.
\end{itemize}
Given a formal grammar, \textit{parsing} is the process of determining whether a string of symbolic nodes can be accepted according to the production rules in the grammar. If the string is accepted by the grammar, the parsing process generates a parse tree. A parse tree represents the syntactic structure of a string according to certain CFG. The root node of the tree is the grammar root. Other non-leaf nodes correspond to non-terminals in the grammar, expanded according to grammar production rules. The leaf nodes are terminal nodes. All the leaf nodes together form a sentence.

In neural-symbolic tasks, the objective of parsing is to find the most probable $z$ that can be accepted by the grammar:
\begin{align} \label{eq:grammar}
    \hat{z} = \arg\max_{z \in L(G)}  p_\theta(z|x)
\end{align}
where $L(G)$ denotes the language of $G$, i.e., the set of all valid $z$ that accepted by $G$.

Traditional grammar parsers can only work on symbolic sentences. \citet{qi2018generalized} proposes a generalized version of Earley Parser, which takes a probability sequence as input and outputs the most probable parse. We use this method to compute the best parse $\hat{z}$ in \autoref{eq:grammar}.

\noindent \textbf{Symbolic Reasoning}. Given the parsed symbolic representation $\hat{z}$, the symbolic reasoning module performs deterministic inference with $\hat{z}$ and the domain-specific knowledge $\Delta$. Formally, we want to find the entailed sentence $\hat{y}$ given $\hat{z}$ and $\Delta$:

\vskip -0.3in
\begin{align}
\hat{y}:\hat{z}~\land~ \Delta \models \hat{y}
\end{align}
\vskip -0.1in

Since the inference process is deterministic, we re-write the above equation as:
\begin{align}
    \hat{y} = f(\hat{z};\Delta),
\end{align}
where $f$ denotes complete inference rules under the domain $\Delta$. The inference rules generate a reasoning path $\hat{\tau}$ that leads to the predicted output $\hat{y}$ from $\hat{z}$ and $\Delta$. The reasoning path $\hat{\tau}$ has a tree structure with the root node $\hat{y}$ and the leaf nodes from $\hat{z}$ or $\Delta$.

\subsection{Learning}
It is challenging to obtain the ground truth of the symbolic representation $z$, and the rules (\ie grammar rules and the symbolic inference rules) are usually designed explicitly by human knowledge. We formulate the learning process as a weakly-supervised learning of the neural network model $\theta$ where the symbolic representation $z$ is missing, and the grammar model $G$, domain-specific language $\Delta$, the symbolic inference rules $f$ are given. 



\subsubsection{1-step back-search ($1$-BS)}
As shown in ~\autoref{fig_RL_vs_BS}, previous methods using policy gradient to learn the model discard all the samples with zero reward and learn nothing from them. It makes the learning process inefficient and unstable. However, humans can learn from the wrong predictions by \textit{diagnosing} and \textit{correcting} the wrong answers according to the desired outputs with top-down reasoning. Based on such observation, we propose a 1-step back-search ($1$-BS) algorithm which can \textit{correct} wrong samples and use the corrections as pseudo labels for training. The $1$-BS algorithm closes the learning loop since the error can also be propagated through the non-differentiable grammar parsing and symbolic reasoning modules. Specifically, we find the most probable correction for the wrong prediction by back-tracking the symbolic reasoning tree and propagating the error from the root node into the leaf nodes in a top-down manner.

The $1$-BS algorithm is implemented with a priority queue as shown in Algorithm~\ref{alg_1_BS}. The $1$-BS gradually searches down the reasoning tree $\hat{\tau}$ starting from the root node $S$ to the leaf nodes. Specifically, each element in the priority queue represents a valid change, defined as a 3-tuple $(A, \alpha_A, p)$:
\begin{itemize}[noitemsep]
    \item $A \in V \cup \Sigma$ is the current visiting node.
    \item $\alpha_A$ is the expected value on this node, which means if the value of $A$ is changed to $\alpha_A$, $\hat{z}$ will execute to the ground-truth answer $y$, \ie $y = f(\hat{z}(A \to \alpha_A);\Delta))$.
    \item $p$ is the visiting priority, which reflects the potential of changing the value of $A$.
\end{itemize}
Formally, the priority for this change is defined as the probability ratio:
\begin{align}
p(A \to \alpha_A) = \left\{
                    \begin{array}{ll}
                    \frac{1-p(A)}{p(A)},&\text{if}~A \notin \Sigma \\
                    \frac{p(\alpha_A)}{p(A)}, &\text{if}~A \in \Sigma~\&~\alpha_A \in \Sigma.
                    \end{array} 
                    \right.
\end{align}
where $p(A)$ is calculated as \autoref{eq_neural},if $A \in \Sigma$; otherwise, it is defined as the product of the probabilities of all leaf nodes in $A$.
If $A \in \Sigma$ and $\alpha_A \notin \Sigma$, it means we need to correct the terminal node to a value that is not in the vocabulary. Therefore, this change is not possible and thus should be discarded.

The error propagation through the reasoning tree is achieved by a $solve(B, A, \alpha_A|\Delta, G)$ function, which aims at computing the expected value $\alpha_B$ of the child node $B$ from the expected value $\alpha_A$ of its parent node $A$, \ie, finding $\alpha_B$ satisfying $f(\hat{z}(B \to \alpha_B);\Delta)) = f(\hat{z}(A \to \alpha_A);\Delta)) = y$. 
 Please refer to the \textit{supplementary material} for some illustrative examples of the $1$-BS process.

In the $1$-BS, we make a greedy assumption that only one symbol can be replaced at a time. This assumption implies only searching the neighborhood of $\hat{z}$ at one-step distance. In ~\autoref{sec:mBS}, the $1$-BS is extended to the multi-step back-search algorithm, which allows searching beyond one-step distance.
\begin{algorithm}[h]
    \caption{$1$-step back-search ($1$-BS)} \label{alg_1_BS}
    \begin{algorithmic}[1]
    \STATE \textbf{Input}: $\hat{z}, S, y$
    \STATE $q = PriorityQueue()$
    \STATE $q.push(S, y, 1)$ 
    \WHILE{$A, \alpha_A, p = q.pop()$}
    	\IF{$A \in \Sigma$}
    	    \STATE $z^* = \hat{z}(A \to \alpha_A)$
    	    \STATE \textbf{return} $z^*$
        \ENDIF
        \FOR{$B \in child(A)$}
            \STATE $\alpha_B = solve(B, A, \alpha_A| \Delta, G)$
            \STATE $q.push(B, \alpha_B, p(B\to\alpha_B))$
        \ENDFOR
    \ENDWHILE
    \STATE \textbf{return} $\varnothing$
    \end{algorithmic}
\end{algorithm}
\vskip -0.7in

\subsubsection{Maximum Likelihood Estimation}
Since $z$ is conditioned on $x$ and $y$ is conditioned on $z$, the likelihood for the observation $(x,y)$ marginalized over $z$ is:
\begin{align} \label{eq:p_yx}
p(y|x) = \sum_z p(y,z|x) = \sum_z p(y|z)p_\theta(z|x).
\end{align}
The learning goal is to maximize the observed-data log likelihood
$L(x,y) = \log p(y|x)$.

By taking derivative, the gradient for the parameter $\theta$ is given by
\begin{align}
\nabla_\theta L(x,y) &=  \nabla_\theta \log p(y|x) \nonumber \\
&= \frac{1}{p(y|x)} \nabla_\theta p(y|x) \nonumber \\
&= \sum_z \frac{p(y|z)  p_\theta(z|x) }{\sum_{z'} p(y|z')p_\theta(z'|x)} \nabla_\theta \log p_\theta(z|x) \nonumber \\
&= \E_{z \sim p(z|x,y)} [\nabla_\theta \log p_\theta(z|x) ], \label{eq_g_L}
\end{align}
where $p(z|x,y)$ is the posterior distribution of $z$ given $x,y$. Since $p(y|z)$ is computed by the symbolic reasoning module and can only be 0 or 1, $p(z|x,y)$ can be written as:
\begin{align} 
p(z|x,y) &= \frac{p(y|z)  p_\theta(z|x) }{\sum_{z'} p(y|z')p_\theta(z'|x)} \nonumber \\
&=\left\{
    \begin{array}{ll}
    0,~~~~\text{for}~~~z \not\in Q\\
    \frac{  p_\theta(z|x) }{\sum_{z' \in Q} p_\theta(z'|x)}, ~~~\text{for}~z \in Q
    \end{array} 
\right.\label{eq_posterior}
\end{align}
where $Q = \{z: p(y|z) = 1\} = \{z:  f(z;\Delta)=y\}$ is the set of $z$ that generates $y$. Usually $Q$ is a very small subset of the whole space of $z$.

\autoref{eq_posterior} indicates that $z$ is sampled from the posterior distribution $p(z|x,y)$, which only has non-zero probabilities on $Q$, instead of the whole space of $z$. 
Unfortunately, computing the posterior distribution is not efficient as evaluating the normalizing constant for this distribution requires summing over all possible $z$, and the computational complexity of the summation grows exponentially. 

Nonetheless, it is feasible to design algorithms that sample from this distribution using Markov chain Monte Carlo (MCMC). Since $z$ is always trapped in the modes where $p(z|x,y) = 0$, the remaining question is how we can sample the posterior distribution $p(z|x,y)$ efficiently to avoid redundant random walk at states with zero probabilities. 


\subsubsection{$m$-BS as Metropolis-Hastings Sampler}
\label{sec:mBS}
\begin{algorithm}[h]
    \caption{$m$-step back-search ($m$-BS)} \label{alg_m_BS}
    \label{alg:training}
    \begin{algorithmic}[1]
    \STATE \textbf{Hyperparameters}: $T$, $\lambda$
    \STATE \textbf{Input}: $\hat{z}, y$
    \STATE $z^{(0)} = \hat{z}$
    \FOR{$t \gets 0~\text{to}~T-1$}
        \STATE $z^* = 1$-BS$(z^{t}, y)$
        \STATE draw $u \sim \mathcal{U}(0,1)$
        \IF{$u \leq \lambda $ and $z^* \neq \varnothing$ }
                \STATE $z^{t+1} = z^*$  
        \ELSE
            \STATE $z^{t+1} = \textsc{RandomWalk}(z^{t})$

        \ENDIF
    \ENDFOR
    \STATE \textbf{return} $z^{T}$
    \STATE
    \FUNCTION{\textsc{RandomWalk}($z^{t}$)}
    		\STATE sample $z^* \sim g(\cdot | z^{t})$ \label{line_for} 
             \STATE compute acceptance ratio $a = min(1, \frac{p_\theta(z^*|x)}{p_\theta(z^{t}|x)})$
              \STATE draw $u \sim \mathcal{U}(0,1)$
              \STATE 
                $
                z^{t+1} = \left\{
                \begin{array}{ll}
                z^*,~&\text{if}~u\leq a\\
                z^{t},~&\text{otherwise}.
                \end{array} 
                \right.
                $
  	\ENDFUNCTION
    
    \end{algorithmic}
    
\end{algorithm}
In order to perform efficient sampling, we extend the 1-step back search to a multi-step back search ($m$-BS), which serves as a Metropolis-Hastings sampler.

A Metropolis-Hastings sampler for a probability distribution $\pi(s)$ is a MCMC algorithm that makes use of a proposal distribution $Q(s'|s)$ from which it draws samples and uses an acceptance/rejection scheme to define a transition kernel with the desired distribution $\pi(s)$. Specifically, given the current state $s$, a sample $s' \neq s$ drawn from $Q(s'|s)$ is accepted as the next state with probability
\begin{align}
    A(s, s') = min \left\{ 1, \frac{\pi(s')Q(s|s')}{\pi(s)Q(s'|s)} \right\}.
\end{align}

Since it is impossible to jump between the states with zero probability, we define $p'(z|x,y)$ as a smoothing of $p(z|x,y)$ by adding a small constant $\epsilon$ to $p(y|z)$:

\begin{align} 
p'(z|x,y) &= \frac{[p(y|z)+\epsilon]  p_\theta(z|x) }{\sum_{z'} [p(y|z')+\epsilon] p_\theta(z'|x)} 
\end{align}

As shown in Algorithm~\ref{alg_m_BS}, in each step, the $m$-BS proposes $1$-BS search with probability of $\lambda$ ($\lambda < 1$) and random walk with probability of $1-\lambda$. The combination of $1$-BS and random walk helps the sampler to traverse all the states with non-zero probabilities and ensures the Markov chain to be ergodic.

\textbf{Random Walk}: Defining a Poisson distribution for the random walk as
\begin{align}
    g(z_1 | z_2) = Poisson(d(z_1, z_2);\beta),
\end{align}
where $d(z_1, z_2)$ denotes the edit distance between $z_1, z_2$, and $\beta$ is equal to the expected value of $d$ and also to its variance. $\beta$ is set as 1 in most cases due to the preference for a short-distance random walk. The acceptance ratio for sampling a $z^*$ from $g(\cdot | z^{t})$ is $ a = min(1, r(z^{t}, z^*))$, where
\begin{align}
    r(z^{t}, z^*) &= \frac{q(z^*) (1-\lambda) g(z^{t} | z^*)}{q(z^{t}) (1-\lambda)g(z^* | z^{t})} \nonumber \\
    &= \frac{p_\theta(z^*|x)}{p_\theta(z^{t}|x)}.
\end{align}

\textbf{$1$-BS}: While proposing the $z^*$ with $1$-BS, we search a $z^*$ that satisfies $p(y|z^*) = 1$. If $z^*$ is proposed, the acceptance ratio for is $a = min(1, r(z^{t}, z^*))$, where
\begin{align}
    r(z^{(t)}, z^*) &= \frac{q(z^*) [ 0 + (1-\lambda)g(z^{t} | z^*)]}{q(z^{t}) \cdot [\lambda + (1-\lambda) g(z^*| z^{(t)})]}  \\
    &= \frac{1 + \epsilon}{\epsilon} \cdot \frac{p_\theta(z^*|x)}{p_\theta(z^{t}|x)} \cdot \frac{(1-\lambda)g(z^{t} | z^*)}{\lambda + (1-\lambda) g(z^* | z^{t})}. \nonumber 
\end{align}
$q(z) = [p(y|z)+\epsilon]  p_\theta(z|x)$ is denoted as the numerator of $p'(z|x,y)$.
With an enough small $\epsilon$, $\frac{1 + \epsilon}{\epsilon} \gg 1$, $r(z^{t}, z^*) > 1$, we will always accept $z^*$.

Notably, the $1$-BS algorithm tries to transit the current state into a state where $z^* =$ $1$-BS$(z^t, y)$, making movements in directions of increasing the posterior probability. Similar to the gradient-based MCMCs like Langevin dynamics~\cite{duane1986theory,welling2011bayesian}, this is the main reason that the proposed method can sample the posterior efficiently.


\subsubsection{Comparison with Policy Gradient}
Since grammar parsing and symbolic reasoning are non-differentiable, most of the previous approaches for neural-symbolic learning use policy gradient like REINFORCE to learn the neural network. Treat $p_\theta(z|x) $ as the policy function and the reward given $z,y$ can be written as:
\begin{align}
r(z, y) =
\left\{
\begin{array}{ll}
      0,~\text{if}~f(z;\Delta) \neq y. \\
      1,~\text{if}~f(z;\Delta) = y.
\end{array} 
\right.
\end{align}
\vskip -0.1in
The learning objective is to maximize the expected reward under current policy $p_\theta$:
\begin{align} \label{eq:R}
R(x,y) &= \E_{z \sim p_\theta(z|x))} r(z,y) =\sum_z p_\theta(z|x)  r(z,y).
\end{align}
\vskip -0.3in
Then the gradient for $\theta$ is:
\begin{align} 
\nabla_\theta R(x,y)&= \sum_z r(z,y)  p_\theta(z|x)  \nabla_\theta \log p_\theta(z|x) \nonumber \\
&= \E_{z \sim p_\theta(z|x))} [ r(z,y)\nabla_\theta \log p_\theta(z|x) ].
\end{align}
\vskip -0.2in
We can approximate the expectation using one sample at each time, and then we get the REINFORCE algorithm:
\begin{align} \label{eq_RL_g}
    \nabla_\theta &= r(z,y)\nabla_\theta \log p_\theta(z|x), z \sim  p_\theta(z|x) \nonumber\\
    &= 
    \left\{
    \begin{array}{ll}
          0,~&\text{if}~f(z;\Delta) \neq y. \\
          \nabla_\theta \log p_\theta(z|x),~&\text{if}~f(z;\Delta) = y.
    \end{array} 
    \right.
\end{align}
\vskip -0.1in
\autoref{eq_RL_g} reveals the gradient is non-zero only when the sampled $z$ satisfies $f(z;\Delta) = y$. However, among the whole space of $z$, only a very small portion can generate the desired $y$, which implies that \textit{the REINFORCE will get zero gradients from most of the samples}. This is why the REINFORCE method converges slowly or even fail to converge, as also shown from the experiments in \autoref{sec:experiments}.

\section{Experiments and Results} \label{sec:experiments}

\subsection{Handwritten Formula Recognition}
\subsubsection{Experimental Setup}

\textbf{Task definition}.
The handwritten formula recognition task tries to recognize each mathematical symbol given a raw image of the handwritten formula. We learn this task in a weakly-supervised manner, where raw image of the handwritten formula is given as input data $x$, and the computed results of the formulas is treated as outputs $y$. The symbolic representation $z$ that represent the ground-truth formula composed by individual symbols is hidden. Our task is to predict the formula, which could further be executed to calculate the final result.

\textbf{HWF Dataset}.
We generate the HWF dataset based on the CROHME 2019 Offline Handwritten Formula Recognition Task\footnote{\url{https://www.cs.rit.edu/~crohme2019/task.html}}. First, we extract all symbols from CROHME and only keep ten digits (0$\sim$9) and four basic operators ($+$,$-$,$\times$, $\div$). Then we generate formulas by sampling from a pre-defined grammar that only considers arithmetic operations over single-digit numbers. For each formula, we randomly select symbol images from CROHME. Overall, our dataset contains 10K training formulas and 2K test formulas.


\textbf{Evaluation Metrics}.
We report both the calculation accuracy (\ie whether the calculation of predicted formula yields to the correct result) and the symbol recognition accuracy (\ie whether each symbol is recognized correctly from the image) on the synthetic dataset.

\textbf{Models}. 
In this task, we use LeNet \cite{lecun2015lenet} as the neural perception module to process the handwritten formula. Before feeding into LeNet, the original image of an formula is pre-segmented into a sequence of sub-images, and each sub-image contains only one symbol. The symbolic reasoning module works like a calculator, and each inference step computes the parent value given the values of two child nodes (left/right) and the operator. The $solve(B,A,\alpha_A)$ function in 1-step back-search algorithm works in the following way for mathematical formulas:
\begin{itemize}
    \setlength\itemsep{0em}
    \item If $B$ is $A$'s left or right child, we directly solve the equation $\alpha_B \bigoplus child_{R}(A) = \alpha_A$ or $child_{L}(A) \bigoplus \alpha_B = \alpha_A$ to get $\alpha_B$, where $\bigoplus$ denotes the operator.
    \item If $B$ is an operator node, we try all other operators and check whether the new formula can generate the correct result.
\end{itemize}
We conduct experiments by comparing the following variants of the proposed model:
\vskip -0.3in
\begin{itemize}
\setlength\itemsep{0em}
\item \textbf{NGS-RL}: learning the NGS model with REINFORCE. 
\item \textbf{NGS-MAPO}: learning the NGS model by Memory Augmented Policy Optimization (MAPO)~\cite{Liang2018MemoryAP}, which leverages a memory buffer of rewarding samples to reduce the variance of policy gradient estimates.
\item \textbf{NGS-RL-Pretrain}: NGS-RL with LeNet pre-trained on a small set of fully-supervised data.
\item \textbf{NGS-MAPO-Pretrain}: NGS-MAPO with pre-trained LeNet.
\item \textbf{NGS-m-BS}: learning the NGS model with the proposed m-step back-search algorithm.
\end{itemize}

\subsubsection{Results and Analyses}
\begin{figure}[ht]
	\centering
    \subfigure{
    \includegraphics[width=\columnwidth]{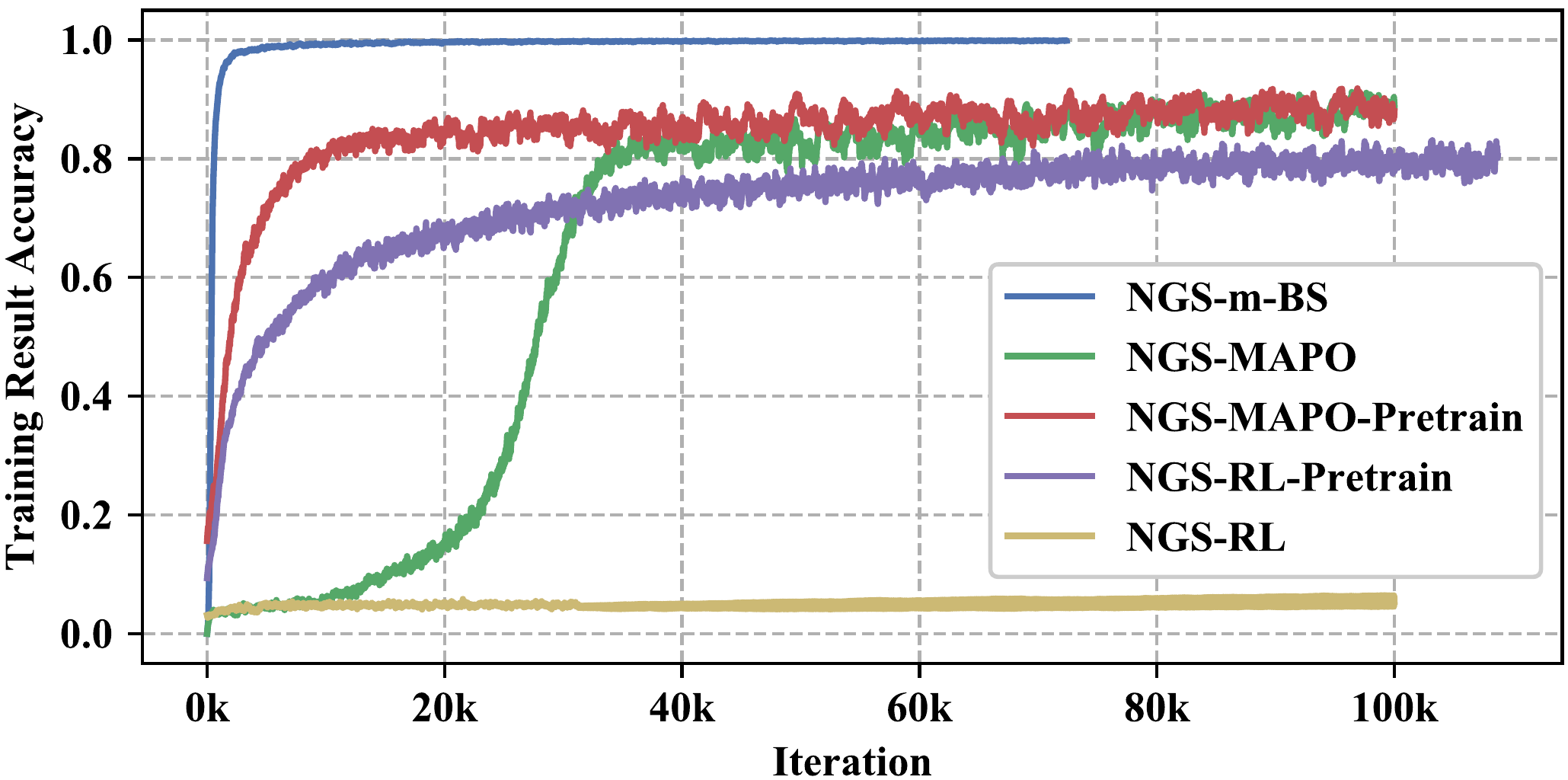}}
    \vskip -0.15in
    \subfigure{
    \includegraphics[width=\columnwidth]{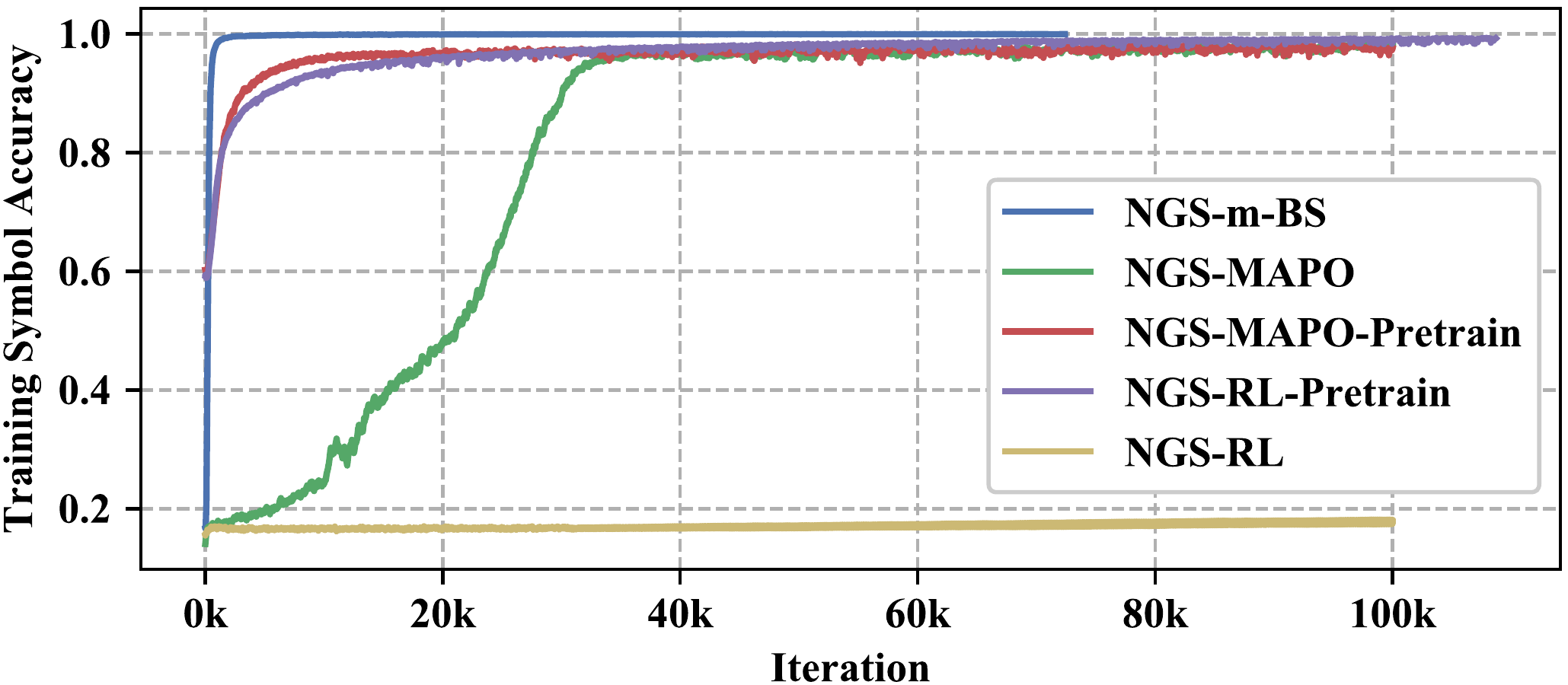}}
    \vskip -0.2in
    \caption{The learning curves of the calculation accuracy and the symbol recognition accuracy for different models.} \label{fig_HME_result_acc}
\end{figure}

\textbf{Learning Curve}. \autoref{fig_HME_result_acc} shows the learning curves of different models. The proposed NGS-m-BS converges much faster and achieves higher accuracy compared with other models. NGS-RL fails without pre-training and rarely improves during the entire training process. NGS-MAPO can learn the model without pre-training, but it takes a long time to start efficient learning, which indicates that MAPO suffers from the cold-start problem and needs time to accumulate rewarding samples. Pre-training the LeNet solves the cold start problem for NGS-RL and NGS-MAPO. However, the training curves for these two models are quite noisy and are hard to converge even after 100k iterations. Our NGS-m-BS model learns from scratch and avoids the cold-start problem. It converges quickly with nearly perfect accuracy, with a much smoother training curve than the RL baselines. 

\begin{figure}[ht]
\centering
\includegraphics[width=\columnwidth]{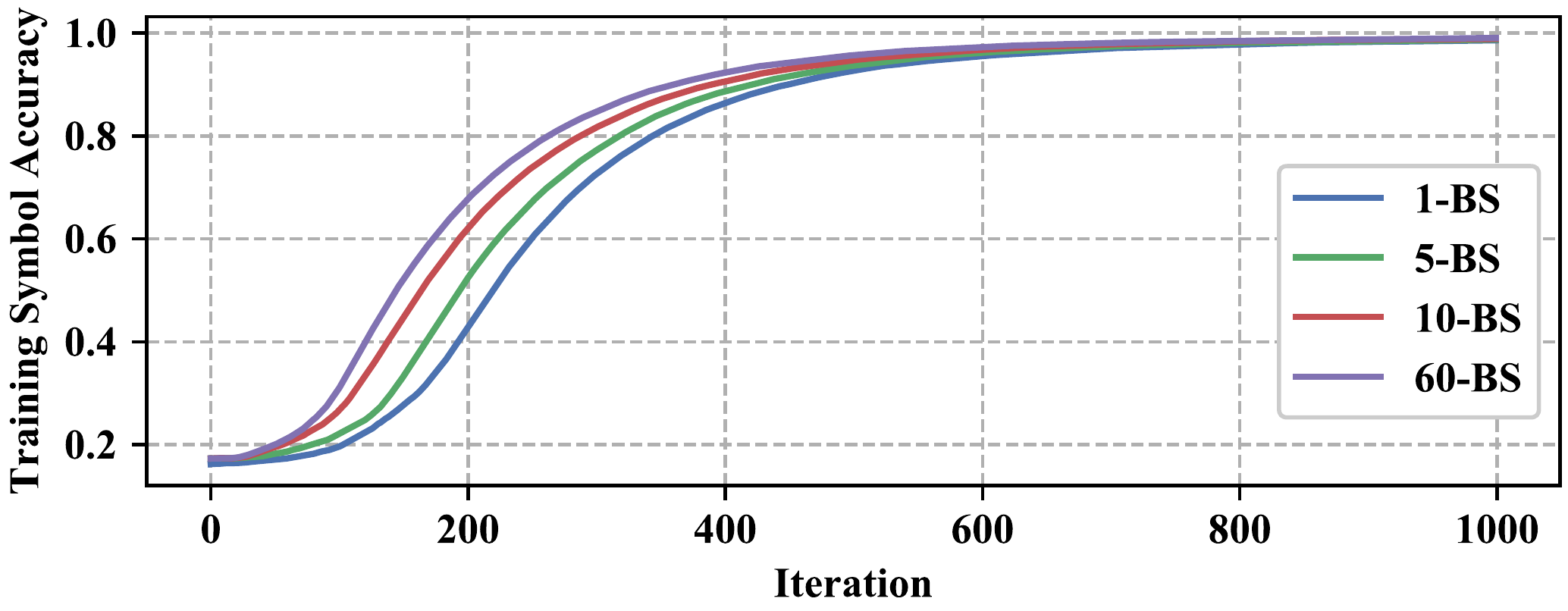} 
\vskip -0.2in
\caption{The training curves of NGS-m-BS with different steps.}
\label{fig_HME_m_BS}
\vskip -0.1in
\end{figure}

\textbf{Back-Search Step}. \autoref{fig_HME_m_BS} illustrates the comparison of the various number of steps in the multi-step back-search algorithm. Generally, increasing the number of steps will increase the chances of correcting wrong samples, thus making the model converge faster. However, increasing the number of steps will also increase the time consumption of each iteration.

\textbf{Data Efficiency}. \autoref{tab_HME_reult_acc_data_percentage} and \autoref{tab_HME_symbol_acc_data_percentage} show the accuracies on the test set while using various percentage of training data. All models are trained with 15K iterations. It turns out the NGS-m-BS is much more data-efficient than the RL methods. Specifically, when only using 25\% of the training data, NGS-m-BS can get a calculation accuracy of 93.3\%, while NGS-MAPO only gets 5.1\%.
\begin{table}[H]
    \centering
    \small
    \caption{The calculation accuracy on the test set using various percentage of training data.}
    \label{tab_HME_reult_acc_data_percentage}
    \begin{tabular}{l|cccc}
    \hline
         {Model} & 25\% & 50 \% & 75 \% & 100\%   \\
         \hline
         {NGS-RL} & 0.035 & 0.036 & 0.034 & 0.034\\
         {NGS-MAPO} & 0.051 & 0.095 & 0.305 & 0.717\\
         {NGS-RL-Pretrain} & 0.534 & 0.621 & 0.663 & 0.685 \\
         {NGS-MAPO-Pretrain} & 0.687 & 0.773 & 0.893 & 0.956\\
         \hline
         {NGS-m-BS} & \textbf{0.933} & \textbf{0.957} & \textbf{0.975} & \textbf{0.985}\\
         \hline
    \end{tabular}
\end{table}

\begin{table}[H]
    \centering
    \small
    \caption{The symbol recognition accuracy on the test set using various percentage of training data.}
    \label{tab_HME_symbol_acc_data_percentage}
    \begin{tabular}{l|cccc}
    \hline
         Model & 25\% & 50 \% & 75 \% & 100\%   \\
         \hline
         NGS-RL & 0.170 & 0.170 & 0.170 & 0.170\\
         NGS-MAPO & 0.316 & 0.481 & 0.785 & 0.967\\
         NGS-RL-Pretrain & 0.916 & 0.945 & 0.959 & 0.964 \\
         NGS-MAPO-Pretrain & 0.962 & 0.983 & 0.985 & 0.991\\
         \hline
         NGS-m-BS & \textbf{0.988} & \textbf{0.992} & \textbf{0.995} & \textbf{0.997}\\
         \hline
    \end{tabular}
    \vskip -0.1in
\end{table}


\begin{figure*}[ht]
\centering
\vskip -0.1in
\includegraphics[width=2\columnwidth]{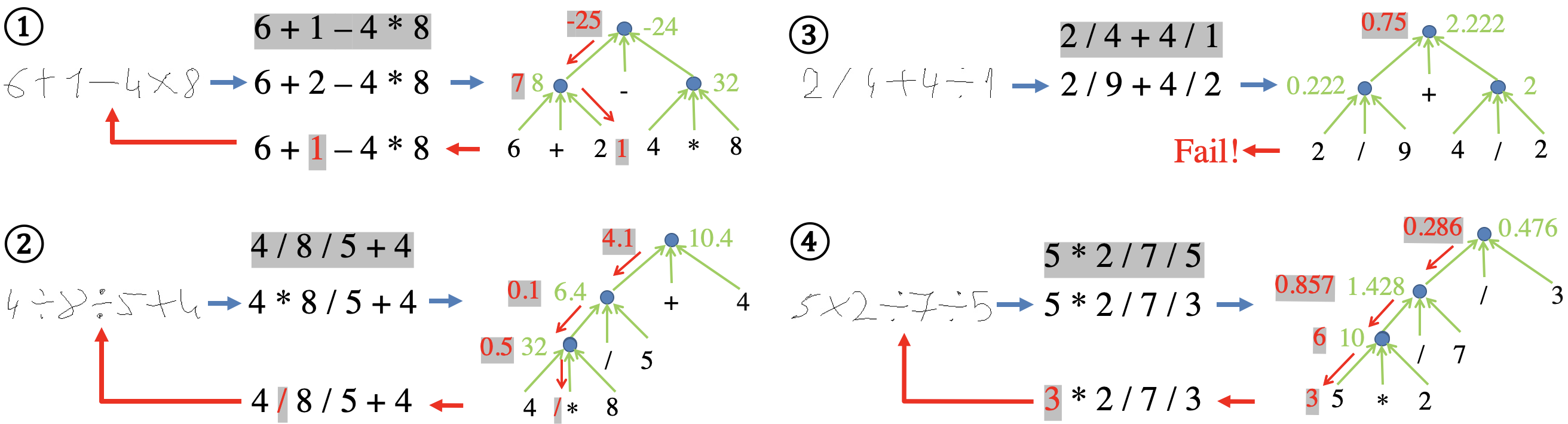} 
\caption{Examples of correcting wrong predictions using the one-step back-search algorithm.} 
\label{fig_HME_examples}
\end{figure*}
\textbf{Qualitative Results}. \autoref{fig_HME_examples} illustrates four examples of correcting the wrong predictions with $1$-BS. In the first two examples, the back-search algorithm successfully corrects the wrong predictions by changing a digit and an operator, respectively. In the third example, the back-search fails to correct the wrong sample. However, if we increase the number of search steps, the model could find a correction for the example. In the fourth example, the back-search finds a spurious correction, which is not the same as the ground-truth formula but generates the same result. Such spurious correction brings a noisy gradient to the neural network update. It remains an open problem for how to avoid similar spurious corrections. 


\subsection{Neural-Symbolic Visual Question Answering}

\subsubsection{Experimental Setup}

\noindent{\textbf{Task}}. Following~\cite{yi2018neural}, the neural-symbolic visual question answering task tries to parse the question into functional program and then use a program
executor that runs the program on the structural scene representation to obtain the answer. The functional program is hidden.

\noindent{\textbf{Dataset}}. We evaluate the proposed method on the CLEVR dataset~\cite{johnson2017clevr}. The CLEVR dataset is a popular benchmark for testing  compositional reasoning capability of VQA models in previous works~\cite{johnson2017inferring,vedantam2019probabilistic}.
CLEVR consists of a training set of 70K images and $\sim$700K questions, and a validation set of 15K images and $\sim$150K questions. We use the VQA accuracy as the evaluation metric.

\noindent{\textbf{Models}}.
We adopt the NS-VQA model in~\cite{yi2018neural} and replace the attention-based seq2seq question parser with a Pointer Network~\cite{Vinyals2015PointerN}. We store a dictionary to map the keywords in each question to the corresponding functional modules. For example, ``red''$\to$``filter color [red]'', ``how many''$\to$ ``count'', and ``what size'' $\to$ ``query size'' \textit{etc}. 
Therefore, the Pointer Network can point to the functional modules that are related to the input question. 
The grammar model ensures that the generated sequence of function modules can form a valid program, which indicates the inputs and outputs of these modules can be strictly matched with their forms. We conduct experiments by comparing following models: \textbf{NS-RL}, \textbf{NGS-RL}, \textbf{NGS-1-BS}, \textbf{NGS-m-BS}.

\subsubsection{Results and Analyses}
\begin{figure}[ht]
	\centering {\includegraphics[width=0.5\textwidth]{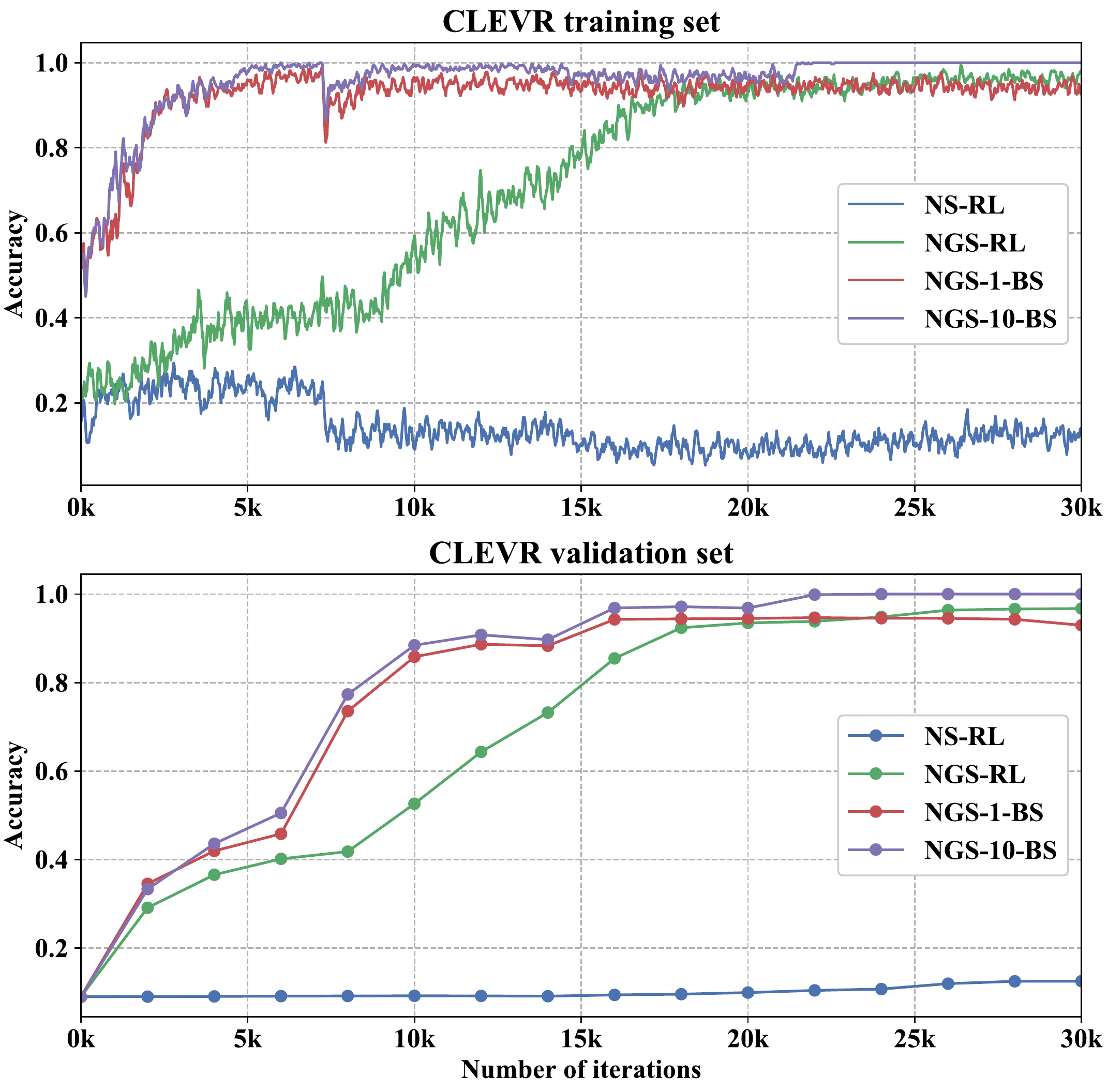}}
	\caption{The learning curve of different model variants on training and validation set of the  CLEVR dataset.}
	\label{fig_vqa_acc}
\end{figure}

\noindent \textbf{Learning Curve}. \autoref{fig_vqa_acc} shows the learning curves of different model variants. NGS-BS converges much faster and achieves higher VQA accuracy on the test set compared with the RL baselines. Though taking a long time, NGS-RL does converge, while NS-RL fails. This fact indicates that the grammar model plays a critical role in this task. Conceivably, the latent functional program space is combinatory, but the grammar model rules out all invalid programs that cannot be executed by the symbolic reasoning module. It largely reduces the solution space in this task.

\noindent \textbf{Back-Search Step}. As shown in \autoref{fig_vqa_acc}, NGS-10-BS performs slightly better than the NGS-1-BS, which indicates that searching multiple steps does not help greatly in this task. One possible reason is that there are more ambiguities and more spurious examples compared with the handwritten formula recognition task, making it less efficient to do the $m$-BS. For example, for the answer ``yes'', there might be many possible programs for this question that can generate the same answer given the image.

\noindent \textbf{Data Efficiency} \autoref{tab_CLEVR_acc_data_percentage} shows the accuracies on the CLEVR validation set when different portions of training data are used. With less training data, the performances decrease for both NGS-RL and NGS-m-BS, but NGS-m-BS still consistently obtains higher accuracies.

\begin{table}[ht]
    \centering
    \caption{The VQA accuracy on the CLEVR validation set using different percentage of training data. All models are trained 30k iterations.}
    \label{tab_CLEVR_acc_data_percentage}
    \begin{tabular}{l|cccc}
    \hline
         {Model} & 25\% & 50 \% & 75 \% & 100\%   \\
         \hline
         {NS-RL} & 0.090 & 0.091 & 0.099 & 0.125\\
         {NGS-RL} & 0.678 & 0.839 & 0.905 & 0.969\\
         \hline
         {NGS-m-BS} & \textbf{0.873} & \textbf{0.936} & \textbf{1.000} & \textbf{1.000}\\
         \hline
    \end{tabular}
\end{table}


\section{Conclusions} \label{sec_conclusion}
In this work, we propose a neural-grammar-symbolic model and a back-search algorithm to close the loop of neural-symbolic learning. We demonstrate that the grammar model can dramatically reduce the solution space by eliminating invalid possibilities in the latent representation space. The back-search algorithm endows the NGS model with the capability of learning from wrong samples, making the learning more stable and efficient. One future direction is to learn the symbolic prior (\ie the grammar rules and symbolic inference rules) automatically from the data.

\paragraph{Acknowledgements.}
We thank Baoxiong Jia for helpful discussion on the generalized Earley Parser. This work reported herein is supported by ARO W911NF1810296, DARPA XAI N66001-17-2-4029, and ONR MURI N00014-16-1-2007.
{\small
\bibliographystyle{icml2020}
\bibliography{main.bib}
}
\clearpage

\end{document}


\icmltitlerunning{Supplementary materials for Neural-Grammar-Symbolic}
\twocolumn[
\icmltitle{Supplementary Materials for \\ Closed Loop Neural-Symbolic Learning via \\ Integrating Neural Perception, Grammar Parsing, and Symbolic Reasoning}
]

\section{REINFORCE as Rejection Sampling}
\begin{align*}
\nabla_\theta L(x,y) &=  \nabla_\theta \log p(y|x) \nonumber \\
&= \frac{1}{p(y|x)} \nabla_\theta p(y|x) \nonumber \\
&= \sum_z \frac{p(y|z)  p_\theta(z|x) }{\sum_{z'} p(y|z')p_\theta(z'|x)} \nabla_\theta \log p_\theta(z|x) \nonumber \\
&= \E_{z \sim p(z|x,y)} [\nabla_\theta \log p_\theta(z|x) ] \\
&\approx \nabla_\theta \log p_\theta(\hat{z}|x), \hat{z} \sim p(z|x,y)
\label{eq_g_L}
\end{align*}

\begin{align*}
    p(z|x,y) \\
    f(\hat{z};\Delta)\neq y \\
    p(\hat{z}|x,y)=0 \\
    \frac{p(\hat{z}|x,y) }{ M p_\theta(\hat{z}|x)} = 1, \text{where } M =\frac{1}{\sum_{z' \in Q} p_\theta(z'|x)}
\end{align*}

\section{Posterior Approximation}
In Section~{3.2.3}, we formulate the $m$-step back search as a Metropolis-Hasting sampler to perform sampling from $p'(z|x,y)$, which is a smoothing of the true posterior distribution $p(z|x,y)$ as shown in Equation~11.
Intuitively, as $\epsilon$ gets smaller, the distance between two distribution $p'(z|x,y)$ and $p(z|x,y)$ becomes smaller as well. Accordingly, we have the following lemma proved with Equation~9 and Equation~11:
\begin{lemma}
Given an small $\epsilon$, the Kullback–Leibler divergence of $p'(z|x,y)$ from $p(z|x,y)$ is $O(\epsilon)$.
\end{lemma}
\begin{proof}
From the definition of Kullback–Leibler divergence, we have:
\begin{align}
KL(p||p') 
&= \sum_z p(z|x,y)\log\frac{p(z|x,y)}{p'(z|x,y)} \\
&= \sum_z p(z|x,y)\log[\frac{p(y|z)}{p(y|z) + \epsilon} \cdot \frac{\epsilon+\sum_{z'}p(y|z')p_{\theta}(z'|x)}{\sum_{z'}p(y|z')p_{\theta}(z'|x)}] \\
&= \sum_{z \in Q} \frac{p_\theta(z|x)}{C}~\log \frac{C+\epsilon}{(1+\epsilon)C} \\
&= \log \frac{C+\epsilon}{(1+\epsilon)C}  \\
&= \log(1+\frac{\epsilon}{C}) - \log{(1+\epsilon)}
\end{align}
where $C = \sum_{z'} p(y|z')p_\theta(z'|x) = \sum_{z' \in Q} p_\theta(z'|x)$ is the normalizing constant.
With Taylor expansion, we get:
\begin{align}
&\log(1+\frac{\epsilon}{C}) = \frac{\epsilon}{C} + O(\epsilon^2) \\
&\log{(1+\epsilon)} = \epsilon + O(\epsilon^2)
\end{align}
Then we have:
\begin{align}
KL(p||p')  = (\frac{1}{C} -1)\epsilon + O(\epsilon^2) = O(\epsilon)
\end{align}
\end{proof}



\section{Handwritten Formula Recognition}

\subsection{Grammar for Math Formulas}

For the handwritten formula recognition task, we define the context-free grammar for the mathematical formulas, as shown in \autoref{tab_math_grammar}. This grammar considers only simple arithmetic operations over single-digit numbers. We compute the parsed results using a calculator, which is the symbolic reasoning module in this task.  

To be noticed, the proposed method can be extended to more complex computations by designing more complicated grammar.

\begin{table}[ht]
\centering
\small
\caption{The context-free grammar for the mathematical formulas.} \label{tab_math_grammar}
\begin{tabular}{l}
\hline
G = (V, $\Sigma$, R, S) \\
\hline
V= \{S, Expression, Term, Factor\} \\
$\Sigma = \{+,-,\times,\div,0,1,...,9 \}$.\\
S is the start symbol. \\
R = \{\\
~~~~~~~~S $\to$ Expression\\
~~~~~~~~Expression $\to$ Term\\
~~~~~~~~Expression $\to$ Expression + Term\\
~~~~~~~~Expression $\to$ Expression - Term\\
~~~~~~~~Term $\to$ Factor\\
~~~~~~~~Term $\to$ Term $\times$ Factor\\
~~~~~~~~Term $\to$ Term $\div$ Factor\\
~~~~~~~~Factor $\to 0 | 1 | 2 | 3 ...| 9$ \}\\
\hline
\end{tabular}
\end{table}

\vspace{-0.1in}
\subsection{Data Generation}
We generate the synthetic dataset based on CROHME 2019 Offline Handwritten Formula Recognition Task\footnote{\url{https://www.cs.rit.edu/~crohme2019/task.html}}. First, we extract all the image patches of symbols from CROHME and only keep ten digits (0$\sim$9) and four basic operators ($+$,$-$,$\times$, $\div$). We split these images of symbols into a training symbol set (80\%) and a testing symbol set (20\%). Then we generate formulas by randomly sampling production rules from the predefined grammar. 
For the training set, we generate 1K formulas with length 1 (1 digit, 0 operator), 1K formulas with length 3 (2 digits, 1 operator), 2K formulas with length 5 (3 digits, 2 operators), and 6K formulas with length 7 (4 digits, 3 operators). 
For the test set, we generate 200 formulas with length 1, 200 formulas with length 3, 400 formulas with length 5, and 1,200 formulas with length 7.
For each formula in the training/test set, we randomly select symbol images from the training/test symbol set.  In this way, one symbol image can not exist in both the training set and the test set. Overall, our dataset contains 10K training formulas and 2K test formulas. The generated dataset is also submitted with the code.

\begin{figure*}[t]
\centering
\includegraphics[width=2\columnwidth]{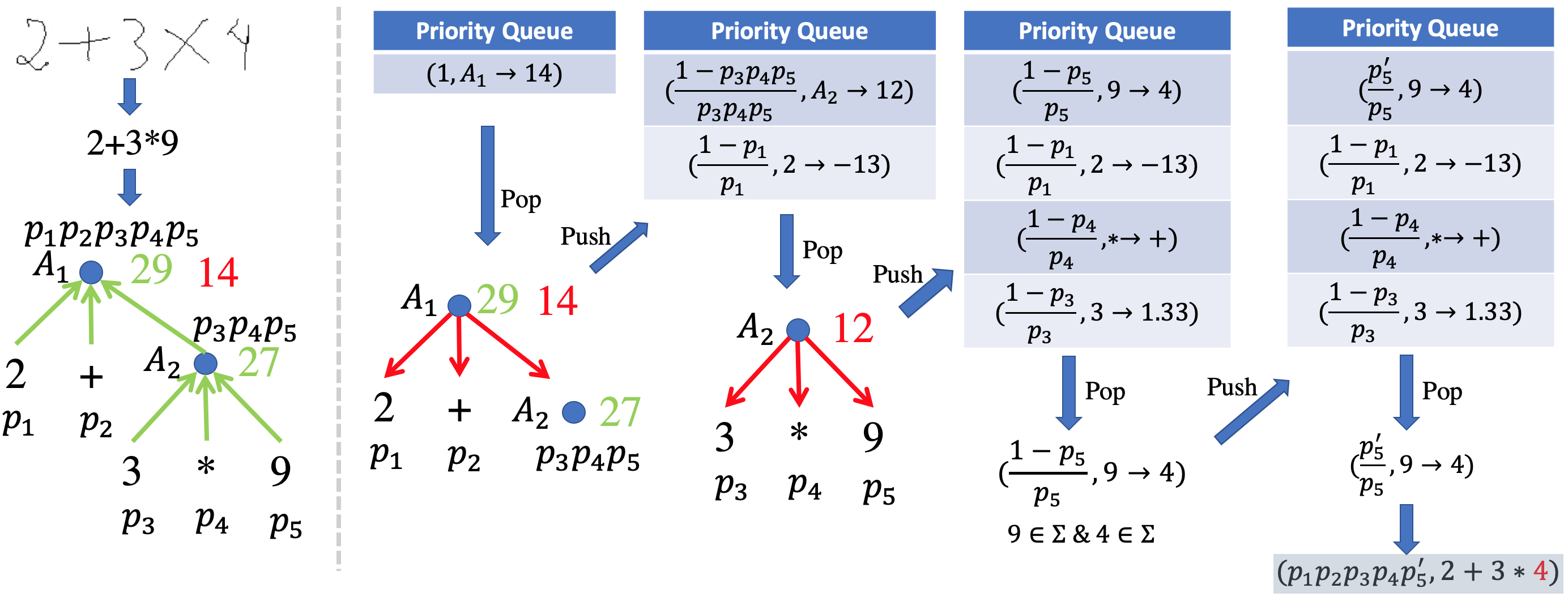}
\vspace{-3mm}
\caption{An illustrative example of the 1-BS process. The priority queue ranks the possible corrections to the original results with visiting priority,
which reflects the potential of changing the current node or
its child nodes to correct the wrong answer.} 
\label{fig_1BS_example}
\end{figure*}

\begin{figure*}[h]
\centering
\includegraphics[width=2\columnwidth]{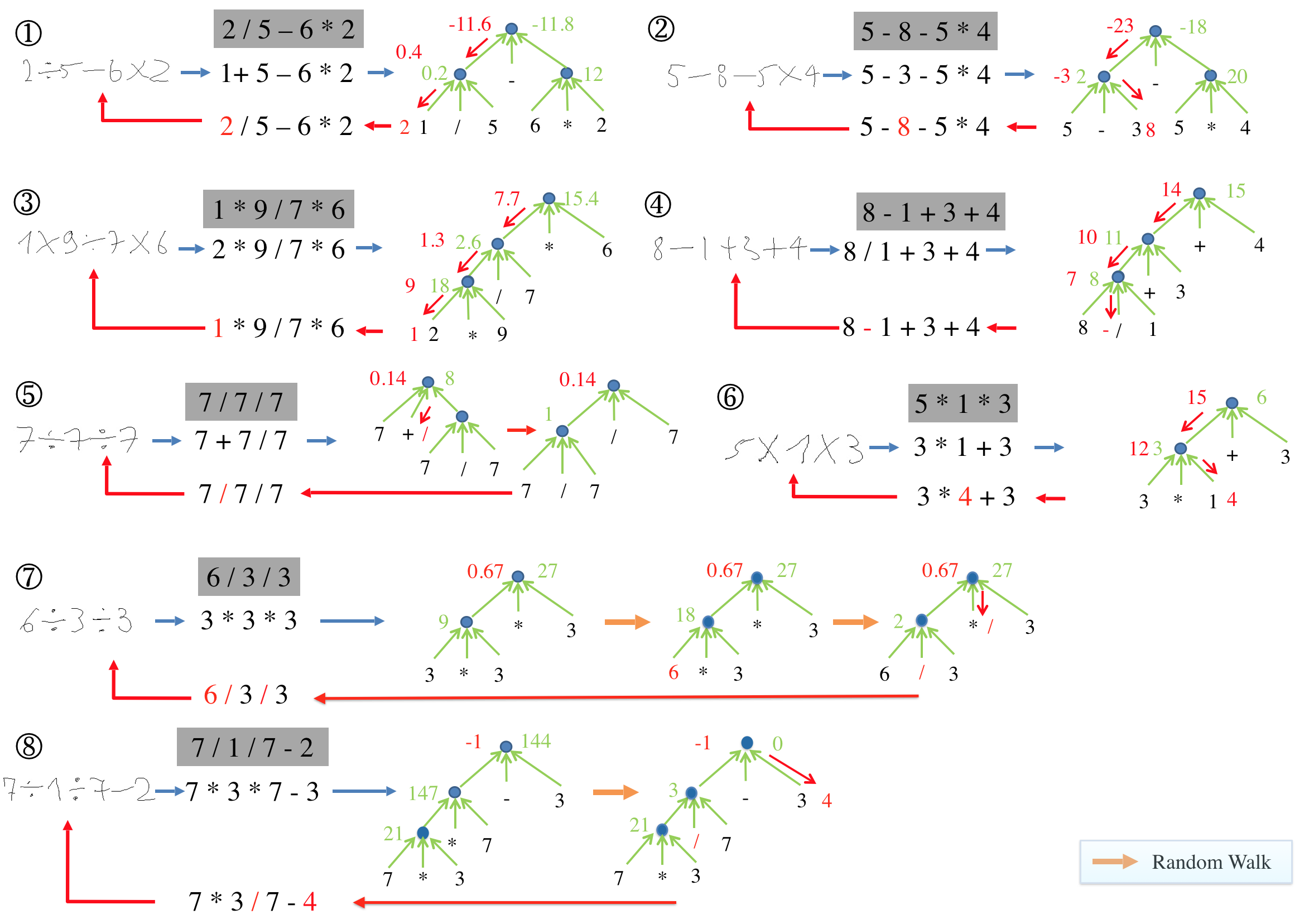} 
\vspace{-3mm}
\caption{Examples of correcting the wrong predictions using the proposed $m$-BS algorithm. Some of the wrong predictions are corrected with the randomly sampled random walks noted by the yellow arrows. (6) and (8) are the spurious examples mentioned in Section 4.1.2} 
\label{fig_mBS_example}
\end{figure*}

\subsection{Training Details}
For the proposed Neural-Grammar-Symbolic models, we use LeNet as the neural perception module and train the models for 100K iterations using the Adam optimizer with a fixed learning rate of $5\times 10^{-4}$ and a batch size of 64. For the REINFORCE and reproduced MAPO baselines, we set the reward decay as $0.99$. For more details in the implementation and reproduction of the experiment results, please refer to the submitted code.

\subsection{Qualitative Examples}
\autoref{fig_1BS_example} shows an illustrative example of the $1$-BS process implemented with a priority queue. \autoref{fig_mBS_example} shows several examples of correcting the wrong predictions using the $m$-BS algorithms.

\section{Neural-Symbolic VQA}
\subsection{Grammar for CLEVR programs}
The grammar model in the neural-symbolic VQA task ensures the generated sequence of function modules can form a valid program, which indicates the inputs and outputs of these modules can be strictly matched. \autoref{tab_CLEVR_modules} groups all function modules by the inputs and output types and \autoref{tab_CLEVR_grammar} gives the context-free grammar for the CLEVR programs.

\begin{table}[h]
\centering
\small
\caption{The context-free grammar for the CLEVR programs.} \label{tab_CLEVR_grammar}
\begin{tabular}{l}
\hline
G = (V, $\Sigma$, R, S) \\
\hline
V = \{S, ObjectSet, Concept, Integer, Object\} \\
$\Sigma$ is the set of all modules as listed in \autoref{tab_CLEVR_modules}.\\
S is the start symbol.\\
R = \{\\
~~~~~~~~S $\to$ \texttt{count} ObjectSet \\
~~~~~~~~S $\to$ \texttt{equal\_attribute} Concept Concept \\
~~~~~~~~S $\to$ \texttt{exist} ObjectSet \\
~~~~~~~~S $\to$ \texttt{greater\_than} Integer Integer \\
~~~~~~~~S $\to$ \texttt{less\_than} Integer Integer \\
~~~~~~~~S $\to$ \texttt{equal\_integer} Integer Integer \\
~~~~~~~~S $\to$ \texttt{query\_attribute} Object \\

~~~~~~~~ObjectSet $\to$ \texttt{scene} \\
~~~~~~~~ObjectSet $\to$ \texttt{filter\_attribute[concept]} ObjectSet \\
~~~~~~~~ObjectSet $\to$ \texttt{intersection} ObjectSet ObjectSet \\
~~~~~~~~ObjectSet $\to$ \texttt{union} ObjectSet ObjectSet \\
~~~~~~~~ObjectSet $\to$ \texttt{relate[RelConcept]} Object \\
~~~~~~~~ObjectSet $\to$ \texttt{same\_attribute} Object \\

~~~~~~~~Concept $\to$ \texttt{query\_attribute} Object \\
~~~~~~~~Integer $\to$ \texttt{count} ObjectSet \\
~~~~~~~~Object $\to$ \texttt{unique} ObjectSet \} \\
\hline
\end{tabular}
\end{table}

\begin{table*}[ht]
    \centering
    \small
    \caption{Modules in the CLEVR programs. They are grouped by their inputs and output signatures. Modules listed in the same group (row) can replace each other while keeping the program valid.}
    \label{tab_CLEVR_modules}
    \resizebox{\textwidth}{!}{
    \begin{tabular}{llll}
    \hline
    \textbf{Modules} & \textbf{Inputs} & \textbf{Output} &  \textbf{Semantics}  \\
    \hline
    \texttt{scene} & $\varnothing$ & ObjectSet &  Return all objects in the scene. \\
    \texttt{count} & ObjectSet & Integer & Count the number of objects in the input object set. \\
    \texttt{equal\_attribute} & Concept, Concept & Bool & Check if two input concepts equal. \\
    \texttt{exist} & ObjectSet & Bool & Check if the object set is empty. \\
    \texttt{filter\_attribute[concept]} & ObjectSet & ObjectSet & Filter out a set of objects having the object-level concept \\
    \texttt{intersection}, \texttt{union} & ObjectSet, ObjectSet & ObjectSet & Return the intersection or union of two object sets.  \\
    \texttt{greater\_than}, \texttt{less\_than}, \texttt{equal\_integer} & Integer, Integer & Bool & Compare two integers. \\
    \texttt{query\_attribute} & Object & Concept & Query the attribute (e.g., color) of the input object. \\
    \texttt{relate[RelConcept]}, \texttt{same\_attribute} & Object & ObjectSet &  Filter out objects with the relational concept or same attribute. \\
    \texttt{unique} & ObjectSet & Object & Return the unique object in the object set. \\
    \hline
    \end{tabular}
    }
\end{table*}

\subsection{Implementation Details}
\begin{figure*}[ht]
\centering
\includegraphics[width=2\columnwidth]{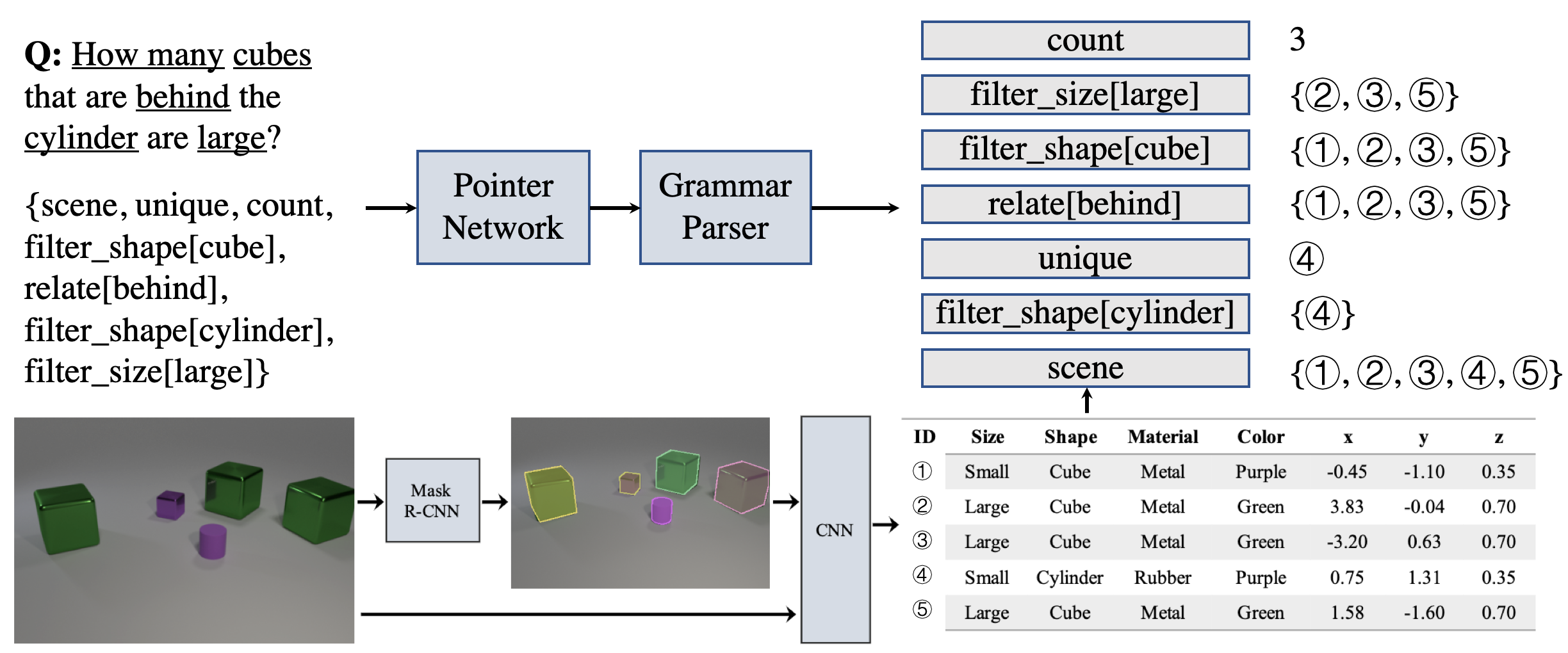} 
\caption{The Neural-Grammar-Symbolic VQA model for the CLEVR dataset.} 
\label{fig_NGS_VQA}
\end{figure*}

The structure of the NGS model is shown in \autoref{fig_NGS_VQA}. To get the structural scene representations, we train a scene parser following~\cite{yi2018neural}. Specifically, Mask-RCNN~\cite{he2017mask} is used to generate segment proposals of all objects in each image. Along with the segmentation mask, the network also predicts the categorical labels of discrete intrinsic attributes such as color, material, size, and shape. The segment for each object is then paired with the original image and sent to a ResNet-34 to extract the spacial attributes such as pose and 3D coordinates. Both networks of the scene parser are trained on 4,000 generated CLEVR images with full annotations. Please refer to~\cite{yi2018neural} for more training details of the scene parser.

Instead of the attention-based seq2seq model used by~\cite{yi2018neural}, we use a Pointer Network as the question parser. Considering the small vocabulary of the CLEVR questions, we can easily build a dictionary to map the keywords in the question to the corresponding modules. Therefore, for each question, we can extract a set of functional modules, and the ground-truth program is a permutation of this set of modules. In the Pointer Network, both the encoder and decoder are two-layer LSTMs with 256 hidden units. We set the dimensions of both the encoder and decoder word embedding to 300. The Pointer Network works as the neural perception module in the proposed NGS model. Unlike~\cite{yi2018neural}, we do not need to pre-train the question parser on a small set of ground-truth question-program pairs.

The symbolic reasoning module in this task executes the generated program on the structural scene representations. The program executor is implemented as a collection of deterministic, generic functions in Python, designed to host all the functional modules in the CLEVR programs. Each function is in one-to-one correspondence with a module from the input program sequence, which has the same representation as in~\cite{johnson2017inferring,yi2018neural}. The execution of a program tree starts from the leaf nodes with \texttt{scene} tokens and continues until the root node, which outputs the final answer to the question.

Since the set of the functional modules is given for each question, the 1-step back search algorithm works by \textit{switching} two modules that belong to the same group according to \autoref{tab_CLEVR_modules}.

All models are trained with 30K iterations using the Adam optimizer with a fixed learning rate of $1\times 10^{-5}$ and a batch size of 64. For the REINFORCE and MAPO baselines, we set the reward decay as $0.99$.

\subsection{Qualitative Examples}
\begin{figure*}[ht]
\centering
\includegraphics[width=2\columnwidth]{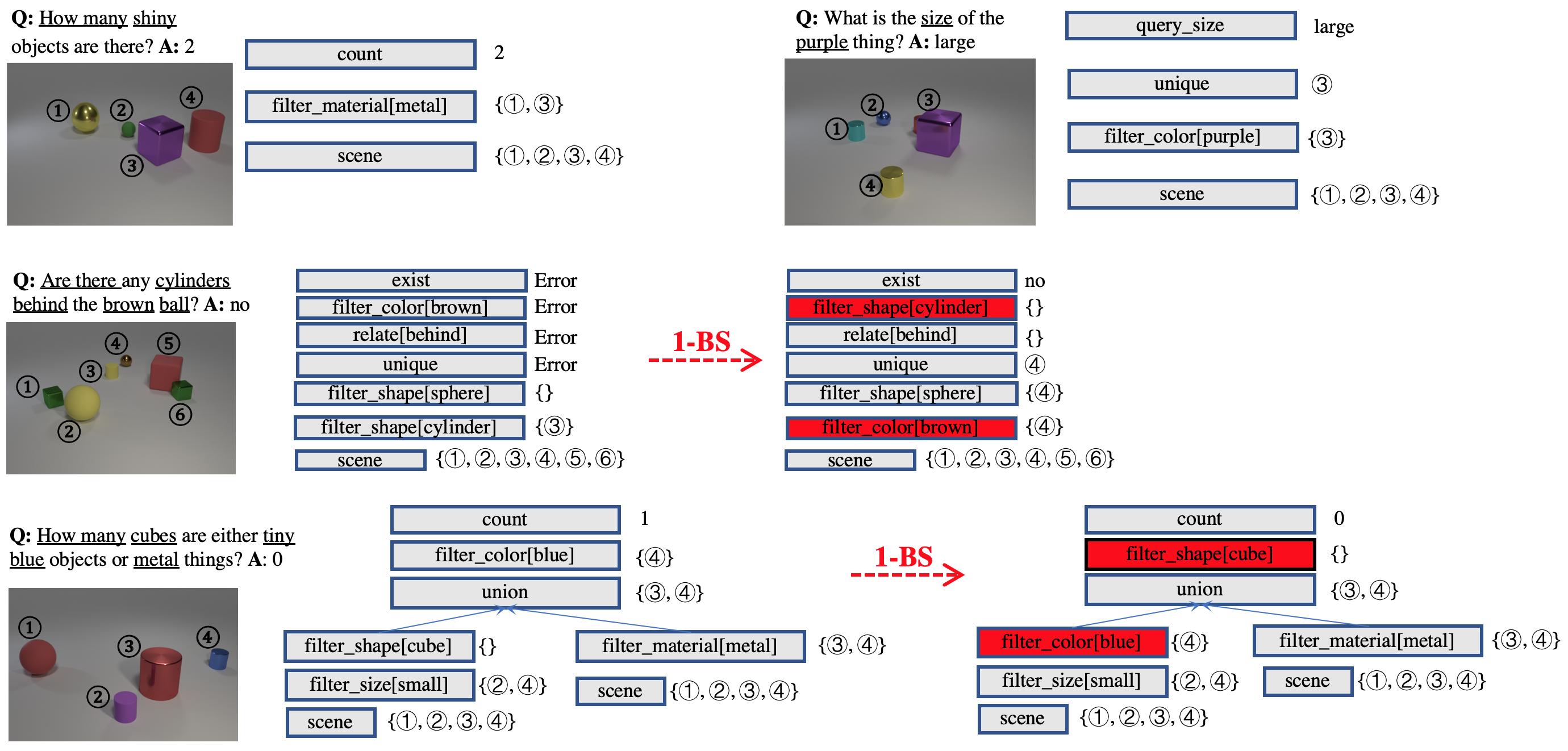} 
\caption{Illustrative examples of correcting the wrong programs using the $1$-BS algorithm. \colorbox[rgb]{ 1,  0,  0}{$***$} denotes the switched modules in the $1$-BS algorithm. In the first two simple examples, given the set of the functional modules, only one permutation can form a valid program, and we do not need to use the back search algorithm. In another two examples, $1$-BS successfully finds the correct programs.}
\label{fig_BS_CLEVR_example}
\end{figure*}

\autoref{fig_BS_CLEVR_example} shows several illustrative examples of correcting the wrong programs using the $1$-BS model. 

{\small
\bibliographystyle{icml2020}
\bibliography{icml.bib}
}
\clearpage